\documentclass{article}

\PassOptionsToPackage{numbers, compress}{natbib}

\usepackage[final]{ML4MH_workshop_neurips2020}

\usepackage[utf8]{inputenc} 
\usepackage[T1]{fontenc}    
\usepackage{hyperref}       
\usepackage{url}            
\usepackage{booktabs}       
\usepackage{amsfonts, amsmath}       
\usepackage{nicefrac}       
\usepackage{microtype}      
\usepackage{tikz}
\usepackage{subcaption}
\usetikzlibrary{bayesnet}
\DeclareMathOperator{\atantwo}{atan2}
\newcommand{\distas}[1]{\mathbin{\overset{#1}{\kern\z@\sim}}}%

\title{Representing and Denoising Wearable \\ ECG Recordings}

\begin{document}

\author{
    Jeffrey Chan\thanks{Work done during an internship at Apple.}\\
    UC Berkeley \\
    \texttt{chanjed@berkeley.edu} \\
    \And
    Andrew C.~Miller \\
    Apple \\
    \texttt{acmiller@apple.com} \\
    \And
    Emily B.~Fox \\
    Apple \\
    \texttt{emily\_fox@apple.com} \\
 }
\date{\today}

\maketitle
\begin{abstract}
Modern wearable devices are embedded with a range of noninvasive biomarker sensors that hold promise for improving detection and treatment of disease. One such sensor is the single-lead electrocardiogram (ECG) which measures electrical signals in the heart. 
The benefits of the sheer volume of ECG measurements with rich longitudinal structure made possible by wearables come at the price of potentially noisier measurements compared to clinical ECGs, e.g., due to movement.
In this work, we develop a statistical model to simulate a structured noise process in ECGs derived from a wearable sensor, design a beat-to-beat representation that is conducive for analyzing variation, and devise a factor analysis-based method to denoise the ECG.
We study synthetic data generated using a realistic ECG simulator and a structured noise model.
At varying levels of signal-to-noise, we quantitatively measure an upper bound on performance and compare estimates from linear and non-linear models.
Finally, we apply our method to a set of ECGs collected by wearables in a mobile health study. 
\end{abstract}

\section{Introduction}
Heart disease is the leading cause of death worldwide, causing over 17.9 million deaths annually with over 600{,}000 of those deaths in the United States alone~\citep{who2020, cdc2020}.
A major challenge in combating heart disease is early identification of high risk individuals.
Recently, wearable devices have enabled individuals to passively track biomarkers of health throughout their day-to-day life, as opposed to sporadically in the clinic.
While these devices hold promise for combating heart disease by allowing for rich longitudinal tracking of biomarkers across large populations of patients, recordings may be noisier than their clinically recorded counterparts.

Electrocardiograms (ECGs), which measure the electrical activity in the heart, are one such biomarker for monitoring cardiovascular activity. Clinical ECGs measure the electrical activity across twelve different spatial views (i.e., leads) of the heart. In addition to standard use, clinical ECGs have been analyzed with pattern recognition algorithms (e.g., neural networks) to make a variety of predictions, including arrhythmias \citep{rajpurkar2017cardiologist, ribeiro2020automatic}, measures of heart failure \citep{attia2019screening}, and risk of future adverse events \citep{miller2019comparison}.

Some modern wearables, however, measure the electrical activity at two locations to produce a single lead. 
In an uncontrolled setting, a single-lead ECG may feature artifacts (e.g., due to movement) or more noise compared to clinical ECGs --- in some extreme cases these artifacts render the ECG unrecognizable.  This noise motivates the need for methods to \emph{denoise} these ECGs.

We propose a generative model for temporally-structured artifacts in ECGs and a two-step denoising procedure rooted in factor analysis.
Our approach estimates both the temporal structure of perturbations across ECGs, and the per-ECG noise amplitude.
In a simulation study, we compare performance of simple baselines (e.g., sample averages) to more complex nonlinear methods (e.g., variational autoencoders \citep{kingma2013auto}) at varying signal-to-noise ratios on synthetic data, comparing reconstruction error.
The relative performance of each estimator depends on the signal-to-noise ratio, and our factor analysis approach consistently performs well and is simpler than nonlinear methods. 
Finally, we apply our method to a subset of ECG records from the Apple Heart and Movement Study\footnote{\url{http://www.bwhresearch.org/appleheartandmovementstudy/}} to demonstrate its efficacy and statistical properties.

\section{Data Description}
Electrocardiograms (ECGs) measure the voltage of the electrical activity of the heart via electrodes placed on the skin (e.g., on the wrist for some wearables).
The cardiac muscle cycles through depolarization and repolarization events, resulting in a structured temporal signal.
Deviations can indicate cardiac abnormalities or disease. Each heartbeat consists of 3 main components: (1) \textit{P wave:} depolarization of the atria, (2) \textit{QRS complex:} depolarization of the ventricles, and (3) \textit{T wave:} repolarization of the ventricles. 

\begin{figure}
    \centering
    \includegraphics[width=.98\textwidth]{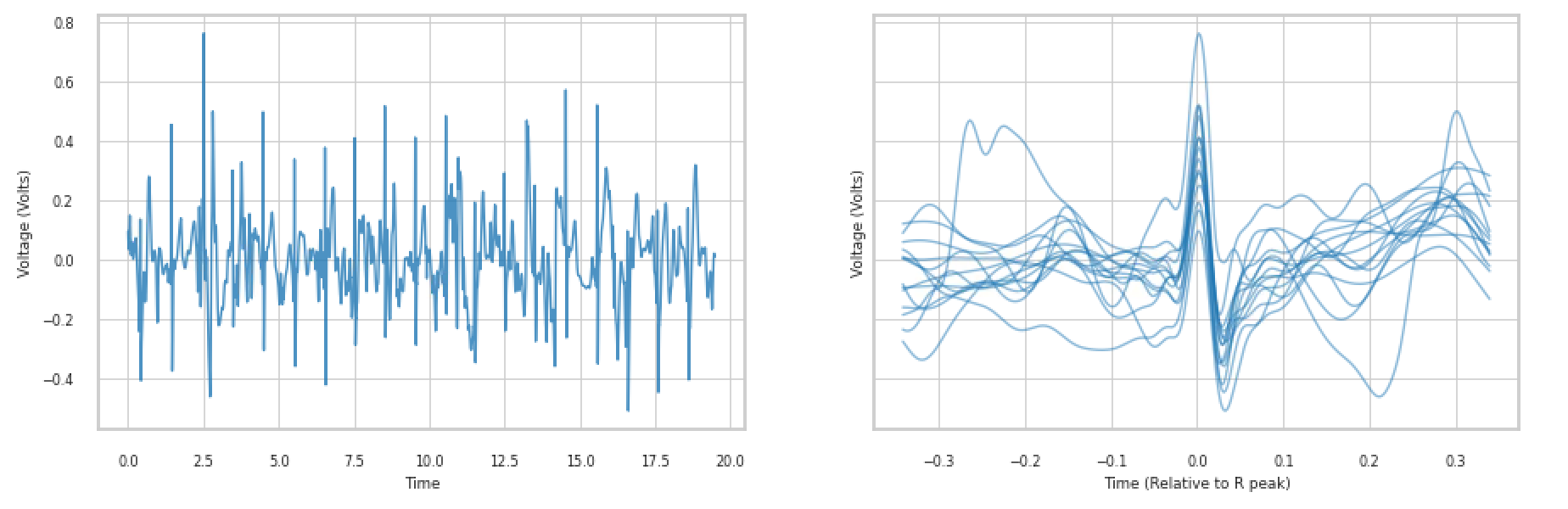}
    \caption{(Left) An example single-lead ECG trace with relatively high noise compared to most measurements. (Right) The same ECG trace represented with the R-peak aligned for each beat.}
    \vspace{-1em}
    \label{fig:representative}
\end{figure}

The data collected by wearables have additional sources of variation (see Figure~\ref{fig:representative}). 
There is heterogeneous noise across samples --- where each sample is a full ECG recording --- typically due to the sensitivity of the wearable to movements and non-ideal conditions.
Cardiac cycle lengths are also heterogeneous, due to changes in heart rhythm either or environmental factors such as exercise.
Additionally, noise is temporally correlated across beats within an ECG sample. 

\subsection{Beat-Aligned Representation}
We first segment each ECG sample into a sequence of beats using a wavelet-based algorithm that delineates the P, QRS, and T waves as proposed in \cite{martinez2004wavelet}.
This delineated representation of the ECG allows us to compare the variability from beat-to-beat within a single ECG trace by aligning each beat $b \in \{1, \ldots, B\}$ relative to its position to the R peak, shown in right of Figure~\ref{fig:representative}. 
For the rest of this work, we assume that the data are collected at a consistent frequency across all ECG samples allowing us to deal with vectors instead of functions, though this can be handled via functional versions of our denoising methods.

\section{Model}
We propose a structured noise model for beat-aligned ECG observations.
Within sample $i$, we model the observed beat $b$ as 
\begin{align}
X_{i,b} = \theta_i + \epsilon_{i,b} \,, && \epsilon_{i,b} \sim N\Big(0, \frac{1}{\tau_i^2} K\Big)\,,
\end{align}
where $X_{i,b}, \theta_i, \epsilon_{i,b} \in \mathbb{R}^d$ represent the observed ECG beat, denoised ECG beat, and the noise, respectively.
The denoised ECG beat $\theta_i$ represents a \emph{per-sample canonical beat}; structured beat-to-beat variation will be considered in future analysis.
ECG noise exhibits temporal correlation structure with varying levels of amplitude, motivating a shared covariance $K$, where we set $tr(K) = d$ for identifiability. 
In order to have a fixed-vector length for each beat across all samples, we fix $d$ so that parts of the beat between the preceding T and succeeding P wave are clipped. 

The inferential goal is to recover $\theta_i$ for each ECG sample from a dataset of observed ECGs, $\{ X_{i,b} \}_{i,b}$.
The simplest approach is to average the aligned beats --- this is standard practice to remove beat-to-beat variability during exercise stress tests \citep{fletcher2013exercise}.
On the other end of the spectrum, a nonlinear latent variable model (e.g., a variational autoencoder) can learn complex global structure to denoise individual ECG samples, but can require large datasets and be difficult to train.

We propose an estimator in between these two extremes --- a two-step approach that first estimates the covariance structure of the global noise $K$ and per-ECG amplitudes $\tau_i$, and then applies factor analysis to a \emph{rotation} of the observed data.
To estimate $K$, we pool information across all observed beats; to estimate the noise amplitude of $\tau_i$ we leverage the multiple observed beats within each recording $i$.
Given an estimate of $K$ and $\tau_i$, we transform each observation $X_{i,b}$ so that stochastic variation has (approximately) diagonal structure, suitable for standard factor analysis. 
The resulting factor analysis estimate is un-transformed to produce the de-noised ECG beat estimate.
Further details of this estimator are in Appendix~\ref{sec:fa-details}. 

In addition to the average beat and VAE, we evaluate the performance of the oracle Bayes estimator \citep{efron2019bayes}, an idealized (and impractical) estimator that serves as an upper bound on estimation performance.
We also evaluate a mixture of factor analyzers model that extends the two-step factor analysis estimator with a flexible learned prior. 
All estimators are described in detail in Appendix~\ref{sec:appendix-model}. 

\section{Experiments}
The denoising methods were run on two datasets: (1) Simulated data where a known ground truth and known noise structure exists and (2) ECGs from a large study where no ground truth data exists. This setup allows us to analyze the statistical properties of the methods in a controlled simulated setting and then study properties of the real data. 
In the simulated setting, we compare the different estimators using the mean squared error of the estimated beat to the ground truth --- this metric averages over the estimator's ability to reconstruct the amplitude and morphology of the ECG beat. 
\subsection{Simulated Data}
\begin{figure}
  \begin{minipage}[c]{0.32\textwidth}
    \caption{
      Left: The variation of the ECG beats based on the jittered ODE parameters. Right: performance of our methods in the multi-beat setting ($B=20$) with variable noise levels ($\tau \sim \mathrm{Unif}(2,20)$).}
    \label{fig:zphys_traces}
  \end{minipage}
  \begin{minipage}[c]{0.36\textwidth}
    \includegraphics[width=\textwidth]{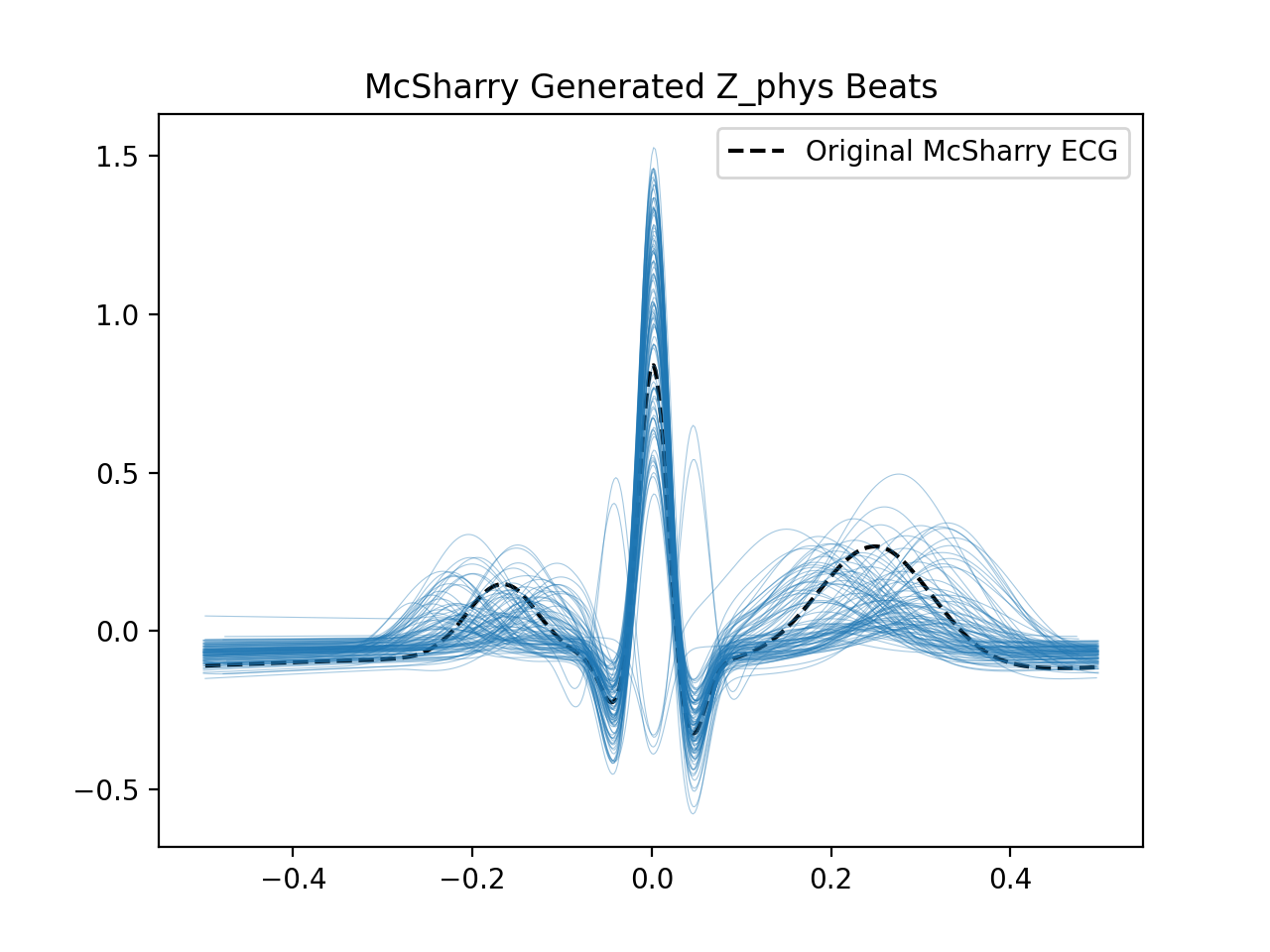}
  \end{minipage}
  \begin{minipage}[l]{0.15\textwidth}
    \scalebox{.8}{
        \begin{tabular}{l|c}
            Method & MSE \\
            \hline
            \hline
            MLE &  0.597\\
            Oracle Bayes, Truth & 0.0\\
            Factor Analysis, Truth & 0.353\\
            Factor Analysis, Estimated & 0.362\\
            VAE, Estimated & 0.660\\
        \end{tabular}
    }
  \end{minipage}
\vspace{-2em}
\end{figure}

The simulated dataset is generated by solving the initial value problem of the following coupled ODE model proposed in \citep{mcsharry2003dynamical} as detailed in Appendix~\ref{sec:appendix-mcsharry}. 
To induce variation in the denoised ECG typically present in real data due to factors such as subject-level physiology or environmental factors, the original parameters in the ECG are jittered and randomly sampled resulting in variation in ECG beats as shown in Figure \ref{fig:zphys_traces}.

We apply the methods described above to the realistic case of $B =20$ and $\tau_i \sim \mathrm{Unif}(2,20)$ for ground-truth and estimated values of $K$ and $\tau$.  The noise covariance $K$ is chosen based on that induced by a Mat\'{e}rn covariance kernel with observations at frequency 500 Hz. 
The oracle Bayes approach is the best case performance we can expect, and we compare our factor analysis-based approaches (linear, post-hoc corrected, and nonlinear) against the standard MLE baseline as shown in Figure \ref{fig:zphys_traces}. The latent space dimension and other hyperparameters for the linear methods were tuned based on the Scree plot where eigenvalues which decayed with a slope of larger than $-0.8$ was chosen as the cutoff. The latent dimension was fixed to $20$ for the VAE.

We find that relative estimator performance depends on noise amplitude.  In the low noise setting, $\tau=20$ with $B = 15$ or $B=20$ beats, it is difficult to outperform the sample average.
However, as noise increases, to $\tau=2$, the factor analysis and nonlinear models significantly outperform the sample average.
While the VAE tends to outperform linear models in the single beat setting, we find that the factor analysis models tend to do better in the multi-beat setting.
Figure~\ref{fig:zphys_traces} compares estimator error when $\tau$ varies uniformly between 2 and 20 and $B=20$; additional results are detailed in Appendix~\ref{sec:additional-results}. 

\begin{figure}[t!]
     \centering
     \begin{subfigure}[b]{0.68\textwidth}
         \centering
         \includegraphics[width=\textwidth]{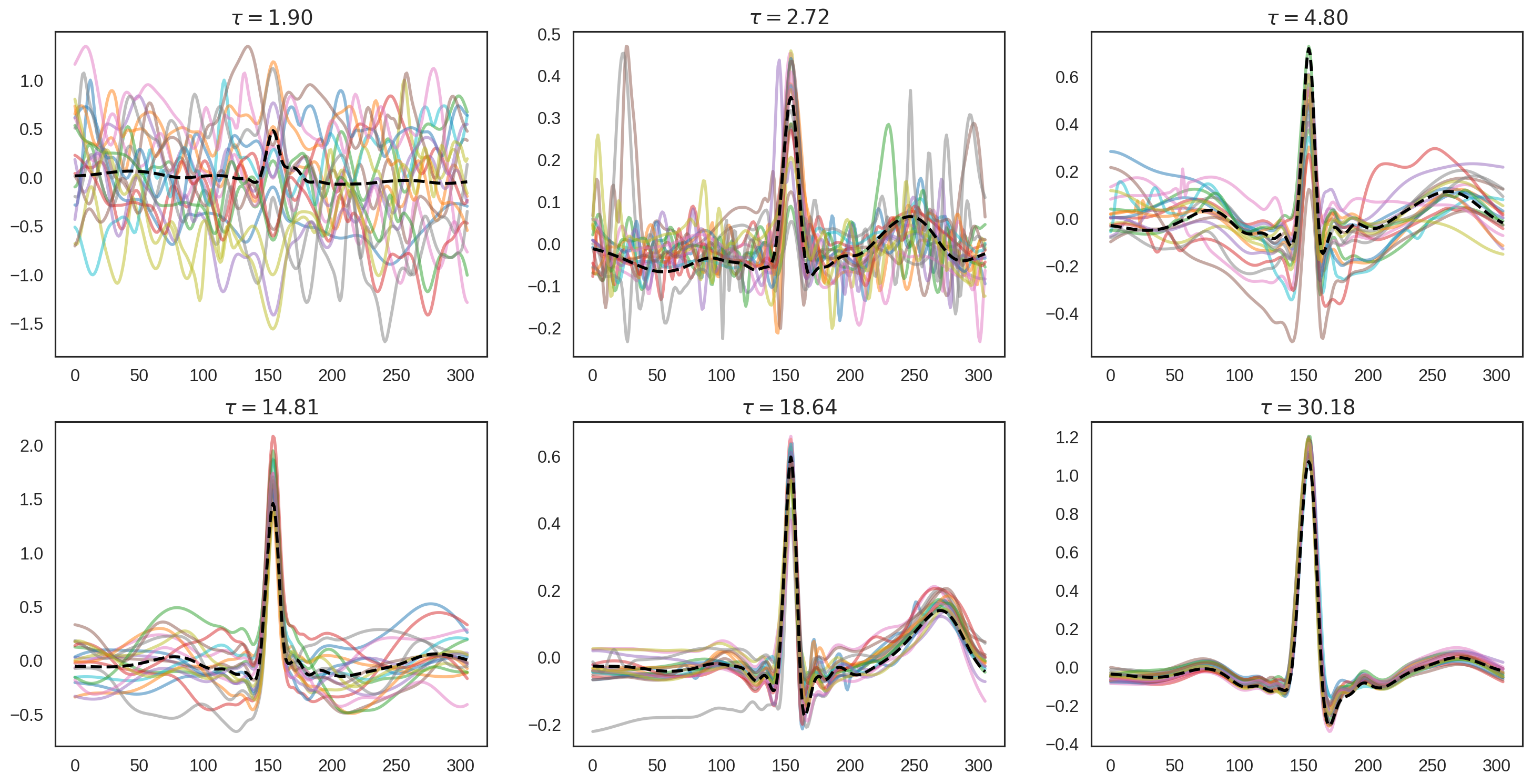}
         \caption{}
         \label{fig:ahms_fa}
     \end{subfigure}
     \hfill
     \begin{subfigure}[b]{0.31\textwidth}
         \centering
         \includegraphics[width=\textwidth]{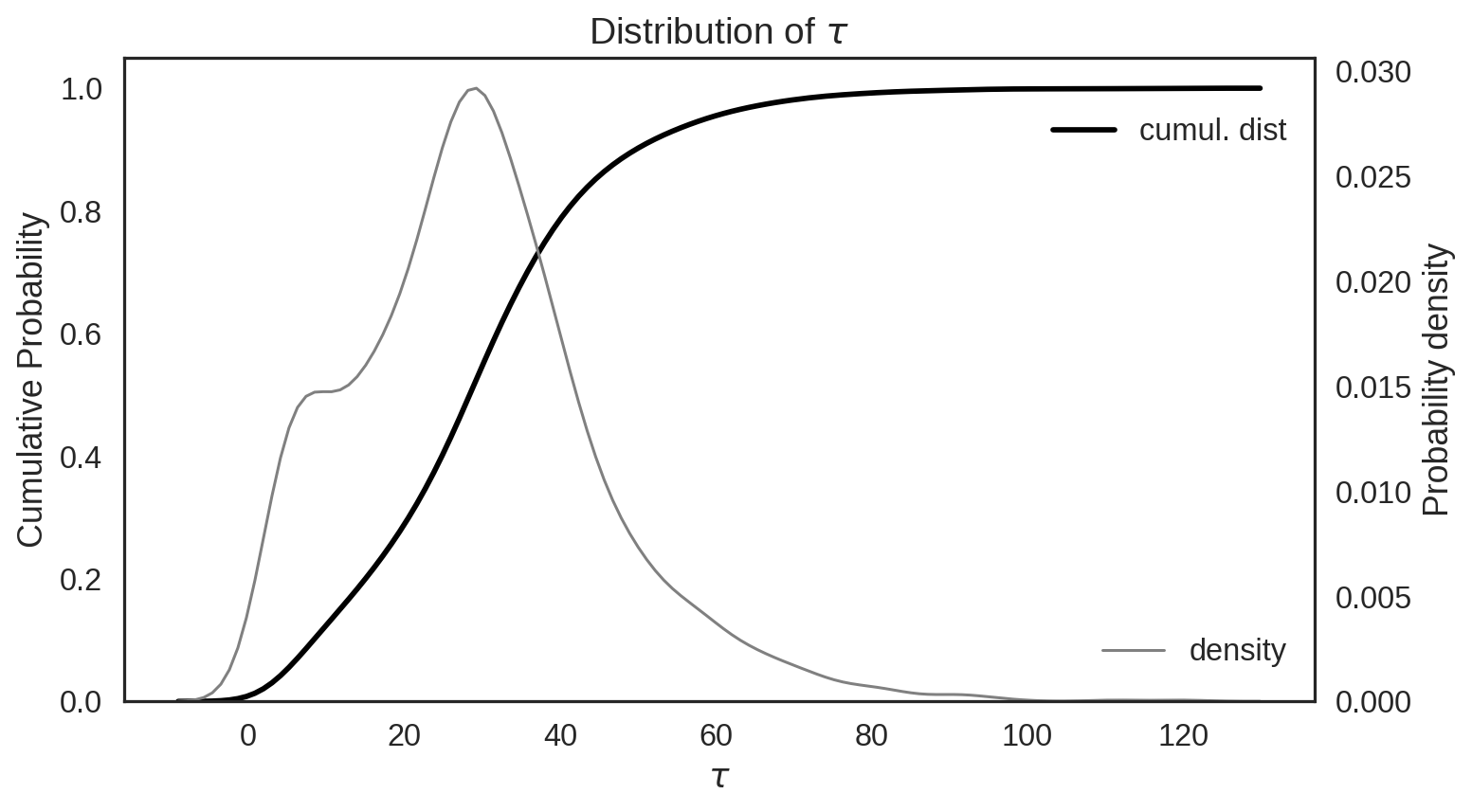} \\
         \includegraphics[width=\textwidth]{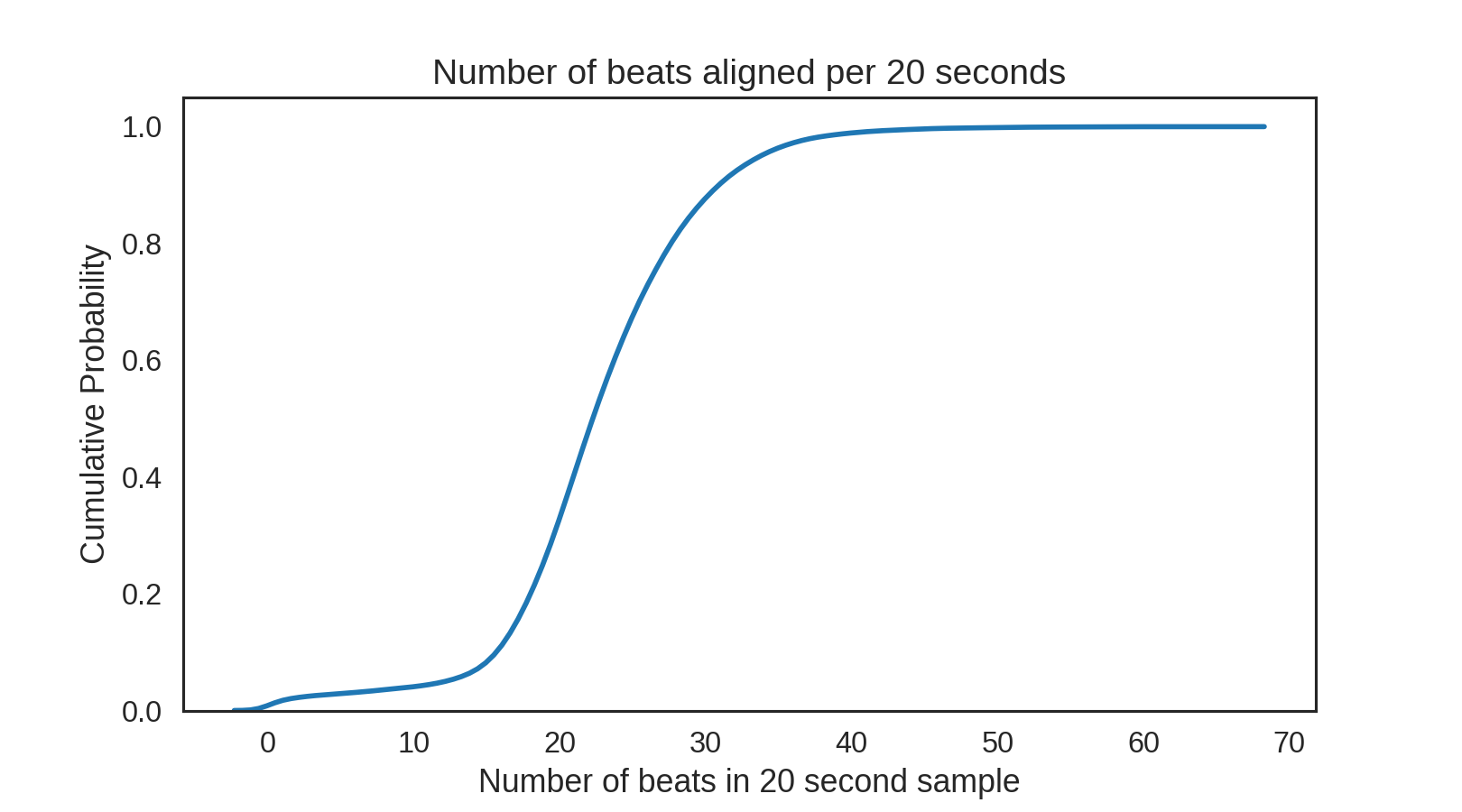}
         \caption{}
         \label{fig:three sin x}
     \end{subfigure}
    \caption{Left: example ECG reconstruction from ECG traces with varying noise levels $\tau$.  The color traces are aligned individual beats, and the dotted line is the factor analysis-reconstructed ECG trace.  Top right: distribution of noise level $\tau$ in a subset of study observed ECGs.  Bottom right: the distribution of the number of beats observed in each recording.}
    \label{fig:ahms-summary}
    \vspace{-1em}
\end{figure}

\subsection{AHMS Data}
The Apple Heart and Movement Study (AHMS) aims to develop better insight into what factors affect heart health over time. Through the Apple Research app, the study collects data from the Apple Watch and iPhone, including ECGs and other mobility signals related to cardiovascular health.\footnote{Information from the Apple Research app is shared with the study only after participants have signed the study informed consent form, and authorized the study to collect or access information in the Apple Research app.  Participant information is encrypted when transferred to and stored on Apple’s servers. Apple is not able to access any information that directly identifies participants (such as name, email address, and phone number) that is collected through the Apple Research app. AHMS was approved by Advarra Institutional Review Board.}

In Figure~\ref{fig:ahms-summary} we show a few representative plots of what our factor analysis approach recovers for participant ECGs of varying noise levels. Also, we plot the distribution of ECGs that could be delineated cleanly. 
For simplicity, we remove any ECG samples with less than $20$ delineated beats.
Among those that passed this delineation filter, we plot the distribution of a subset of ECG noise levels, and find that about 80\% of records have a $\tau > 18$ and over 50\% have a $\tau > 30$.  
As $\tau > 18$ is relatively low noise (see Figure~\ref{fig:ahms_fa}) this indicates that a simple average may suffice for the majority of observations, and a model-based estimate could be used for the remainder. 
Note that the delineation step will under-sample noisier records, which motivates a more robust method for alignment. 

\section{Discussion}
Modeling variation in an ECG waveform is a challenging problem.
ECGs combine a mixture of physiological structure and nuisance variation that are difficult to disentangle. 
We proposed a simple method for representing and denoising a set of ECGs with varying noise-levels.
In a simulation study, we find that in some noise regimes, this approach can be more accurate than simple averaging and competitive with more complex non-linear representation learning methods.
In this work, we do not address rhythmic variation (e.g., heart rate variability) or physiological beat-to-beat variation in ECG morphology (e.g., T wave alternans).  Approaches to disentangle beat-to-beat artifacts from beat-to-beat physiology require additional structural assumptions and methodology, which is an important avenue for continued research.

\small
\bibliographystyle{plainnat}
\bibliography{ref}

\begin{thebibliography}{12}
\providecommand{\natexlab}[1]{#1}
\providecommand{\url}[1]{\texttt{#1}}
\expandafter\ifx\csname urlstyle\endcsname\relax
  \providecommand{\doi}[1]{doi: #1}\else
  \providecommand{\doi}{doi: \begingroup \urlstyle{rm}\Url}\fi

\bibitem[Attia et~al.(2019)Attia, Kapa, Lopez-Jimenez, McKie, Ladewig, Satam,
  Pellikka, Enriquez-Sarano, Noseworthy, Munger, et~al.]{attia2019screening}
Zachi~I Attia, Suraj Kapa, Francisco Lopez-Jimenez, Paul~M McKie, Dorothy~J
  Ladewig, Gaurav Satam, Patricia~A Pellikka, Maurice Enriquez-Sarano, Peter~A
  Noseworthy, Thomas~M Munger, et~al.
\newblock Screening for cardiac contractile dysfunction using an artificial
  intelligence--enabled electrocardiogram.
\newblock \emph{Nature medicine}, 25\penalty0 (1):\penalty0 70--74, 2019.

\bibitem[Chan et~al.(2018)Chan, Perrone, Spence, Jenkins, Mathieson, and
  Song]{chan2018likelihood}
Jeffrey Chan, Valerio Perrone, Jeffrey Spence, Paul Jenkins, Sara Mathieson,
  and Yun Song.
\newblock A likelihood-free inference framework for population genetic data
  using exchangeable neural networks.
\newblock In \emph{Advances in neural information processing systems}, pages
  8594--8605, 2018.

\bibitem[Efron(2019)]{efron2019bayes}
Bradley Efron.
\newblock {B}ayes, oracle {B}ayes and empirical {B}ayes.
\newblock \emph{Statistical Science}, 34\penalty0 (2):\penalty0 177--201, 2019.

\bibitem[Fletcher et~al.(2013)Fletcher, Ades, Kligfield, Arena, Balady,
  Bittner, Coke, Fleg, Forman, Gerber, et~al.]{fletcher2013exercise}
Gerald~F Fletcher, Philip~A Ades, Paul Kligfield, Ross Arena, Gary~J Balady,
  Vera~A Bittner, Lola~A Coke, Jerome~L Fleg, Daniel~E Forman, Thomas~C Gerber,
  et~al.
\newblock Exercise standards for testing and training: a scientific statement
  from the american heart association.
\newblock \emph{Circulation}, 128\penalty0 (8):\penalty0 873--934, 2013.

\bibitem[for Disease~Control and Prevention(2020 (accessed September 15,
  2020)]{cdc2020}
Centers for Disease~Control and Prevention.
\newblock Heart disease facts, 2020 (accessed September 15, 2020.
\newblock URL \url{{https://www.cdc.gov/heartdisease/facts.htm}}.

\bibitem[Kingma and Welling(2014)]{kingma2013auto}
Diederik~P Kingma and Max Welling.
\newblock Auto-encoding variational {B}ayes.
\newblock \emph{International Conference on Learning Representations}, 2014.

\bibitem[Mart{\'\i}nez et~al.(2004)Mart{\'\i}nez, Almeida, Olmos, Rocha, and
  Laguna]{martinez2004wavelet}
Juan~Pablo Mart{\'\i}nez, Rute Almeida, Salvador Olmos, Ana~Paula Rocha, and
  Pablo Laguna.
\newblock A wavelet-based {ECG} delineator: evaluation on standard databases.
\newblock \emph{IEEE Transactions on biomedical engineering}, 51\penalty0
  (4):\penalty0 570--581, 2004.

\bibitem[McSharry et~al.(2003)McSharry, Clifford, Tarassenko, and
  Smith]{mcsharry2003dynamical}
Patrick~E McSharry, Gari~D Clifford, Lionel Tarassenko, and Leonard~A Smith.
\newblock A dynamical model for generating synthetic electrocardiogram signals.
\newblock \emph{IEEE transactions on biomedical engineering}, 50\penalty0
  (3):\penalty0 289--294, 2003.

\bibitem[Miller et~al.(2019)Miller, Obermeyer, and
  Mullainathan]{miller2019comparison}
Andrew~C Miller, Ziad Obermeyer, and Sendhil Mullainathan.
\newblock A comparison of patient history-and {EKG}-based cardiac risk scores.
\newblock \emph{AMIA Summits on Translational Science Proceedings},
  2019:\penalty0 82, 2019.

\bibitem[Organization(2017 (accessed September 15, 2020)]{who2020}
World~Health Organization.
\newblock Cardiovascular diseases, 2017 (accessed September 15, 2020.
\newblock URL
  \url{{https://www.who.int/health-topics/cardiovascular-diseases}}.

\bibitem[Rajpurkar et~al.(2017)Rajpurkar, Hannun, Haghpanahi, Bourn, and
  Ng]{rajpurkar2017cardiologist}
Pranav Rajpurkar, Awni~Y Hannun, Masoumeh Haghpanahi, Codie Bourn, and Andrew~Y
  Ng.
\newblock Cardiologist-level arrhythmia detection with convolutional neural
  networks.
\newblock \emph{arXiv preprint arXiv:1707.01836}, 2017.

\bibitem[Ribeiro et~al.(2020)Ribeiro, Ribeiro, Paix{\~a}o, Oliveira, Gomes,
  Canazart, Ferreira, Andersson, Macfarlane, Wagner~Jr,
  et~al.]{ribeiro2020automatic}
Ant{\^o}nio~H Ribeiro, Manoel~Horta Ribeiro, Gabriela~MM Paix{\~a}o, Derick~M
  Oliveira, Paulo~R Gomes, J{\'e}ssica~A Canazart, Milton~PS Ferreira, Carl~R
  Andersson, Peter~W Macfarlane, Meira Wagner~Jr, et~al.
\newblock Automatic diagnosis of the 12-lead {ECG} using a deep neural network.
\newblock \emph{Nature communications}, 11\penalty0 (1):\penalty0 1--9, 2020.

\end{thebibliography}

\newpage
\appendix
\section*{Appendix}

\section{Beat-Aligned Representation}
\label{sec:beat-aligned-representation}
The single-lead ECG trace is a function $x(t)$ observed for $L$ time steps $t_1 \ldots t_L$ as shown on the left of Figure~\ref{fig:representative}. ECG traces in this time-series representation are difficult to compare due to the cyclical nature of the electrical activity and translation-invariance of the ECG in time. In addition, variability in heart rate results in some nuisance temporal variation of the heartbeats. In order to directly compare the structural features of interest of two ECG traces, we apply a wavelet-based algorithm that delineates the P, QRS, and T waves as proposed in \cite{martinez2004wavelet}. The ECG trace is delineated into the following structure:
\begin{align*}
    \textbf{p} = (p_1, \ldots, p_B) \,, 
    \textbf{q} = (q_1, \ldots, q_B) \,,
    \textbf{r} = (r_1, \ldots, r_B) \,,
    \textbf{s} = (s_1, \ldots, s_B) \,,
    \textbf{t} = (t_1, \ldots, t_B) \,,
\end{align*}
where $p, q, r, s, \text{ and } t$ correspond to the respective timepoints where each wave occurs, and $B$ is the number of beats in the ECG trace. For simplicity, we can assume that the distances between each neighboring wave are consistent within a sample as we find this to be largely the case in the real data we observe later on.\footnote{Subjects with arrhythmias violate this assumption necessitating more detailed temporal models.}

\section{Model}
\label{sec:appendix-model}
\textbf{Assumptions:}
\begin{itemize}
    \item $X$ is the observed ECG beat
    \item $\theta$ is the denoised ECG beat
    \item $\epsilon \sim N(0, (1/\tau)^2 K)$ is the structured noise such that $X = \theta + \epsilon$ and $K$ is fixed and known
    \item $N$ observations
    \item Three cases for $\tau$: (1) $\tau$ is fixed for all observations and known (2) $\tau$ is fixed but unknown and (3) $\tau$ is variable but known.
\end{itemize}
Note that the multi-beat replicate setup allows us to estimate $K$ and $\tau$ for each ECG sample corresponding to the known $K$ and the variable, but known $\tau$ case. Analysis of the single beat case along with estimation of $\tau$ and $K$ allows us to decouple \textit{denoising error} and \textit{noise estimation error}.

\subsection{Maximum Likelihood}
Since $\epsilon$ has zero-mean, the maximum likelihood estimator becomes $$\hat{\theta} = \sum_{b=1}^B X_b \,,$$ 
with the corresponding mean squared error $tr(\frac{1}{\tau^2} K)$ for fixed $\tau$ (case 1 or 2). In the variable, but known $\tau$ (case 3) setting, the MSE becomes
$$\mathrm{MSE} = \frac{1}{N} \sum_{i=1}^N tr\Big(\frac{1}{\tau_i^2} K\Big) \,.$$

\subsection{Oracle Bayes}
Assume we have access to the oracle prior
$$g^{\mathrm{oracle}}(\theta) = \frac{1}{N} \sum_{i=1}^N \delta_{\theta_i}\,.$$
Computing the corresponding posterior yields
\begin{align*}
p(\theta_i|X) &\propto p(X|\theta_i) g^{\mathrm{oracle}}(\theta_i)\\
&\propto N\Bigg(X \Big| \theta_i, \frac{1}{\tau^2} K\Bigg) \,.
\end{align*}

Regardless of whether $\tau$ is fixed or known, the MAP estimate corresponds to the closest $\theta_i$ in the rotated space
$$\hat{\theta}_{MAP} = \arg\min_{\theta \in \theta_1 \ldots \theta_N} \Big\|K^{-1/2}(\theta - X)\Big\|_2\,.$$

\subsection{Factor Analysis}
\label{sec:fa-details}
The factor analysis model modified to incorporate structured noise rather than diagonal noise gives us:
\begin{align*}
    z &\sim N(0, I_p) \\
    \epsilon &\sim N(0, \frac{1}{\tau^2} K)\\
    x - \mu &= Lz + \epsilon \,,
\end{align*}
with the loadings $L \in \mathbb{R}^{d \times p}$, the factors $Z$, and $\mu$ is the mean. We transform the data such that:
\begin{align*}
\tilde{x}^{(i)} &= K^{-1/2}(x^{(i)} - \mu)\\
\tilde{\epsilon} &\sim N(0, \frac{1}{\tau^2}I_d)\\
\tilde{L} &= K^{-1/2}L\\
\tilde{x} &= \tilde{L}z + \tilde{\epsilon} \,.
\end{align*}
This induces a prior over $\theta$ of the form: $g^{\text{FA}}(\theta) = N(\theta | \mu, LL^T)$.
The likelihood function for $L$ can be written as 
\begin{align*}
l(L) &= p(x|L) = \int_z N(x|Lz, \frac{1}{\tau^2} K) N(z| 0, I_p) dz\\
&= N(x| 0, LL^T + \frac{1}{\tau^2}K) \\
&= N(K^{-1/2}x\big|0, (K^{-1/2}L)(K^{-1/2}L)^T + \frac{1}{\tau^2}I_d) \,.
\end{align*}
This coincides with standard factor analysis on the transformed data $K^{-1/2}(X - \mu)$ where the desired loadings can be computed from the recovered loadings $L'$ via $L = K^{1/2}L$. The factor analysis posterior mean is then:
\begin{align*}
E_{p(z^{(i)}|x^{(i)}}[z] &= L^T(LL^T + D)^{-1}(\tilde{x}^{(i)}) \\
\hat{\theta}_{FA} &= K^{1/2}LL^T(LL^T + D)^{-1}K^{-1/2}(x^{(i)} - \mu) \,.
\end{align*}

\subsection{Empirical Bayes Mixture of Gaussian Factor Analysis}
The factor analysis model is appealing in its simplicity. However, the Gaussian prior may be too inflexible to bridge the gap between Factor Analysis and Oracle Bayes. Instead, we propose fitting a Gaussian Mixture Model prior inferred from the data in an empirical Bayes fashion then performing Factor Analysis with this empirical Bayes prior. We start by transforming the data to fit the standard Factor Analysis setup rather than our modified setup with structured noise:
$$\tilde{x}^{(i)} = K^{-1/2}(x^{(i)} - \mu)\,.$$
Then we fit the parameters $L_{EB}$ and $D_{EB}$ to $\tilde{x}$ resulting in the posterior mean of each datapoint 
$$\tilde{z}^{(i)} = \mathbb{E}_{p(z|x^{(i)})}[z] \,.$$
Then, we can fit the set of $\{\tilde{z}\}$ to a Gaussian mixture prior resulting in components $\{(\pi_c, \mu_c, \Sigma_c)\}_{c=1}^C$. We then re-fit our new Factor Mixture Analysis Model
\begin{align*}
    c &\sim \mathrm{Cat}(\pi_1 \ldots \pi_C) \\
    z &\sim N(\mu_c, \Sigma_c)\\
    \epsilon &\sim N(0, D) \\
    \tilde{x} &= Lz + \epsilon \,.
\end{align*}
In order to compute the posterior $p(z|\tilde{x})$, we have:
\begin{align*}
    p(z,x) &= \int_c p(x, z|c) p(c) dc\\
    &= \sum_{c=1}^C \pi_c p(x,z|c) \,.
\end{align*}
Note that $p(z,\tilde{x}|c)$ is jointly Gaussian with the following parameters:
\begin{align*}
\begin{pmatrix}
z\\
\tilde{x}\\
\end{pmatrix}
&\sim
N
\begin{pmatrix}
\begin{bmatrix}
\mu_c \\
L\mu_c
\end{bmatrix}\!,&
\begin{bmatrix}
\Sigma_c & (L\Sigma_c)^T \\
L\Sigma_c & L\Sigma_cL^T + D
\end{bmatrix}
\end{pmatrix} \,.
\end{align*}
We fit $L_{FA}$ and $D_{FA}$ via EM with the following E-step:
\begin{align*}
E[z|x,c] &= \mu_c + (L\Sigma_c)^T(L\Sigma_cL^T + D)^{-1}(x - L\mu_c)\\
V[z|x,c] &= \Sigma_c - (L\Sigma_c)^T (L\Sigma_cL^T + D)^{-1} L\Sigma_c \,.
\end{align*}
In order to compute the posterior mean we have:
\begin{align*}
    E[z|x] &= \sum_c \pi_c \Bigg[\mu_c + (L\Sigma_c)^T(L\Sigma_cL^T + D)^{-1}(x - L\mu_c) \Bigg] \\
    \hat{\theta}_{EB-FA} &= K^{1/2}L E[z|x] \,.
\end{align*}

\subsection{Variational Autoencoder}
The variational autoencoder allows us to devise a more flexible empirical Bayes prior over $\theta$.
\begin{align*}
    z &\sim \mathrm{Normal}(\mu, \sigma^2I_d)\\
    \theta &= \mathrm{Decoder}(z) \\
    X_b &\overset{\text{iid}}{\sim} \mathrm{Normal}\Big(\theta, \frac{1}{\tau^2}K\Big) \,.
\end{align*}
The decoder induces an empirical Bayes prior over $\theta$ that is more flexible than the Normal prior. Inference is done via amortized variational inference. The decoder is designed to be permutation-invariant as done in \cite{chan2018likelihood} in order to allow for a variable number of beats to be used.
\subsection{Multi-beat Estimation of $\tau$ and $K$}
Suppose now that $\tau_i$ and $K$ are unobserved where each $X_{i,b} \sim N(\theta, K/\tau_i^2)$. When $b > 1$, we can estimate $K$ and $\tau_i$. $\tau_i$ represents the relative levels of noise between ECG samples. Define $C_i = K/\tau_i^2$ and $\sigma_i = 1/\tau_i$ and $S = \sum_i \sigma_i^2$. Then, we estimate the $i$-th covariance matrix as 
$$\hat{C}_i = \frac{1}{n-1} \sum_b X_{i,b} X_{i,b}^T$$
Now in order to estimate $S$, we have:
\begin{align*}
\hat{S} &= tr\Big(\sum_i \hat{C}_i\Big) \\ 
\hat{K} &= \frac{d}{\hat{S}}\sum_i \hat{C}_i \\
\hat{\sigma}_i^2 &= \frac{1}{\hat{\tau}^2} = tr\Bigg(\hat{C_i}^{1/2}\hat{K}^{-1}\hat{C_i}^{1/2}\Bigg) \,.
\end{align*}
\section{McSharry ODE Model}
\label{sec:appendix-mcsharry}
The McSharry coupled ODE model is written as:
\begin{align*}
    \frac{du}{dt} &= \alpha u - \omega v \\
    \frac{dv}{dt} &= \alpha v + \omega u \\
    \frac{dx}{dt} &=  - \Bigg[\sum_{i \in \{P, Q, R, S, T\}} a_i \Delta \theta_i \exp\Big(-\frac{\Delta \theta_i^2}{2b_i^2}\Big)\Bigg] - (x - x_0)\\
    \Delta \theta_i &= \atantwo(v,u) - \theta_i \pmod{2\pi} \\
    \alpha &= 1 - \sqrt{u^2 + v^2} \,,
\end{align*}
with corresponding ODE parameters: $a_i, b_i, \theta_i, \text{ and } x_0.$ The $u$ and $v$ variables induce a limit cycle to maintain the consistent structure from beat-to-beat of an ECG. The $x$ variable corresponding to the voltage with respect to time is comprised of a weighted average of 5 Gaussian-like bumps representing each wave in the first term combined with a term that pushes the ECG signal towards the baseline voltage $x_0$. $a_i, b_i, \text{ and } \theta_i$ can be thought of as the weight, bandwidth, and position of each Gaussian bump, respectively.

\section{Additional Results}
\label{sec:additional-results}
\begin{table}[h]
    \centering
    \begin{tabular}{|c||c|c|c|c|c|c|}
    \hline
         &  \multicolumn{6}{c|}{Single Beat}\\
         \hline
         $\tau$ & 2 & 5 & 10 & 15 & 20 & Uniform \\
         \hline
         \hline
         MLE & 123.21 & 19.71 & 4.92 & 2.19 & 1.23 & 11.88\\
         Oracle Bayes & 0.16 & 0.0 & 0.0 & 0.0 & 0.0 & 0.26\\
         FA-Truth & 2.38 & 1.29 & 1.12 & 1.10 & 0.65 & 1.20\\
         MoG-FA & 2.19 & 1.30 & 1.12 & 1.10 & 0.66 & - \\
         VAE-Truth & 2.37 & 1.15 & 0.82 & 0.71 & 0.65 & 1.06\\
         \hline
    \end{tabular}
    \vspace{.5em}
    \caption{Performance of denoising methods in the single-beat setting $B=1$ with varying levels of noise precision $\tau$.}
    \label{tab:my_label}
\end{table}

\begin{table}[h]
    \centering
    \begin{tabular}{|c||c|c|c|c|c|c|}
    \hline
         &  \multicolumn{6}{c|}{Multi Beat}\\
         \hline
         $\tau$ & 2 & 5 & 10 & 15 & 20 & Uniform \\
         \hline
         \hline
         MLE & 6.21 & 0.99 & 0.25 & 0.11 & 0.06 & 0.59\\
         Oracle Bayes & 0.0 & 0.0 & 0.0 & 0.0 & 0.0 & 0.0\\
         FA-Truth & 1.35 & 0.58 & 0.29 & 0.25 & 0.24 & 0.35\\
         FA-Estimated & 1.31 & 0.58 & 0.29 & 0.25 & 0.24 & 0.36 \\
         VAE-Estimated & 0.78 & 0.66 & 0.64 & 0.63 & 0.61 & 0.66\\
         \hline
    \end{tabular}
    \vspace{.5em}
    \caption{Performance of denoising methods in the multi-beat setting $B=20$ with varying levels of noise precision $\tau$.}
    \label{tab:my_label}
\end{table}
\end{document}